\documentclass[10pt,twocolumn,letterpaper]{article}

\usepackage[pagenumbers]{cvpr} 

\usepackage{graphicx}
\usepackage{amsmath}
\usepackage{amssymb}
\usepackage{booktabs}
\usepackage{enumerate}
\usepackage{paralist}
\usepackage{multirow}
\newtheorem{theorem}{Theorem}

\usepackage{bm}
\usepackage[table]{xcolor}
\usepackage{color,colortab}
\definecolor{lightgray}{gray}{0.9}
\definecolor{LightCyan}{rgb}{0.88,1,1}
\usepackage{soul}
\newcommand{\mb}[1]{\mathbf{#1}}
\newcommand{\bs}[1]{\boldsymbol{#1}}

\usepackage[pagebackref,breaklinks=true,colorlinks,citecolor=blue,urlcolor=blue,linkcolor=blue,bookmarks=false]{hyperref}
\usepackage[capitalize]{cleveref}
\crefname{section}{Sec.}{Secs.}
\Crefname{section}{Section}{Sections}
\Crefname{table}{Table}{Tables}
\crefname{table}{Tab.}{Tabs.}

\def\confName{CVPR}
\def\confYear{2023}

\begin{document}
\title{Progressive Multi-resolution Loss for Crowd Counting }
\author{
    Ziheng Yan$^{1}$ \quad Yuankai Qi$^{2}$ \quad Guorong Li$^{1}$\thanks{Corresponding author.} \quad Xinyan Liu$^{1}$
    \\
    Weigang Zhang$^{3}$ \quad Qingming Huang$^{1,4,5}$ \quad Ming-Hsuan Yang$^{6}$
    \\
    $^{1}$University of Chinese Academy of Science, Beijing, China
    \\
    $^{2}$Australian Institute for Machine Learning, The University of Adelaide
    \\
    $^{3}$Harbin Institute of Technology, Weihai, China,
    \\
    $^{4}$Key Lab of Intell. Info. Process., Inst. of Comput. Tech., CAS, Beijing, China
    \\
    $^{5}$Peng Cheng Laboratory, Shenzhen, China, $^{6}$University of California, Merced
    \\
    \small \{yanziheng21, liuxinyan19\}@mails.ucas.ac.cn, 
    qykshr@gmail.com, 
    \\
    \small
    \{liguorong, qmhuang\}@ucas.ac.cn,
    wgzhang@hit.edu.cn, 
    mhyang@ucmerced.edu
}

\maketitle
 
\begin{abstract}
    Crowd counting is usually handled in a density map regression fashion, which is supervised via a L2 loss between the predicted density map and ground truth.
    To effectively regulate models, various improved L2 loss functions have been proposed to find a better correspondence between predicted density and annotation positions.
    In this paper, we propose to predict the density map at one resolution but measure the density map at multiple resolutions. By maximizing the posterior probability in such a setting, we obtain a log-formed multi-resolution L2-difference loss, where the traditional single-resolution L2 loss is its particular case. 
    We mathematically prove it is superior to a single-resolution L2 loss. 
    Without bells and whistles, the proposed loss substantially improves several baselines  and performs favorably compared to state-of-the-art methods  on four crowd counting datasets, ShanghaiTech A \& B, UCF-QNRF, and JHU-Crowd++.
\end{abstract}

\section{Introduction}
\label{sec:intro}
	
	Crowd counting aims to count the number of humans in a crowd scene, which has drawn increasing attention in recent years due to its wide applications in the real world. 
	Most existing methods generate a density map from the input image and then match it to the corresponding dot annotations. 
	One significant challenge in crowd counting is to associate the pixels of the predicted density map to the target dot annotations in a reasonable way. 
	\begin{figure}[tp]
		\centering
		\includegraphics[width=0.77\linewidth]{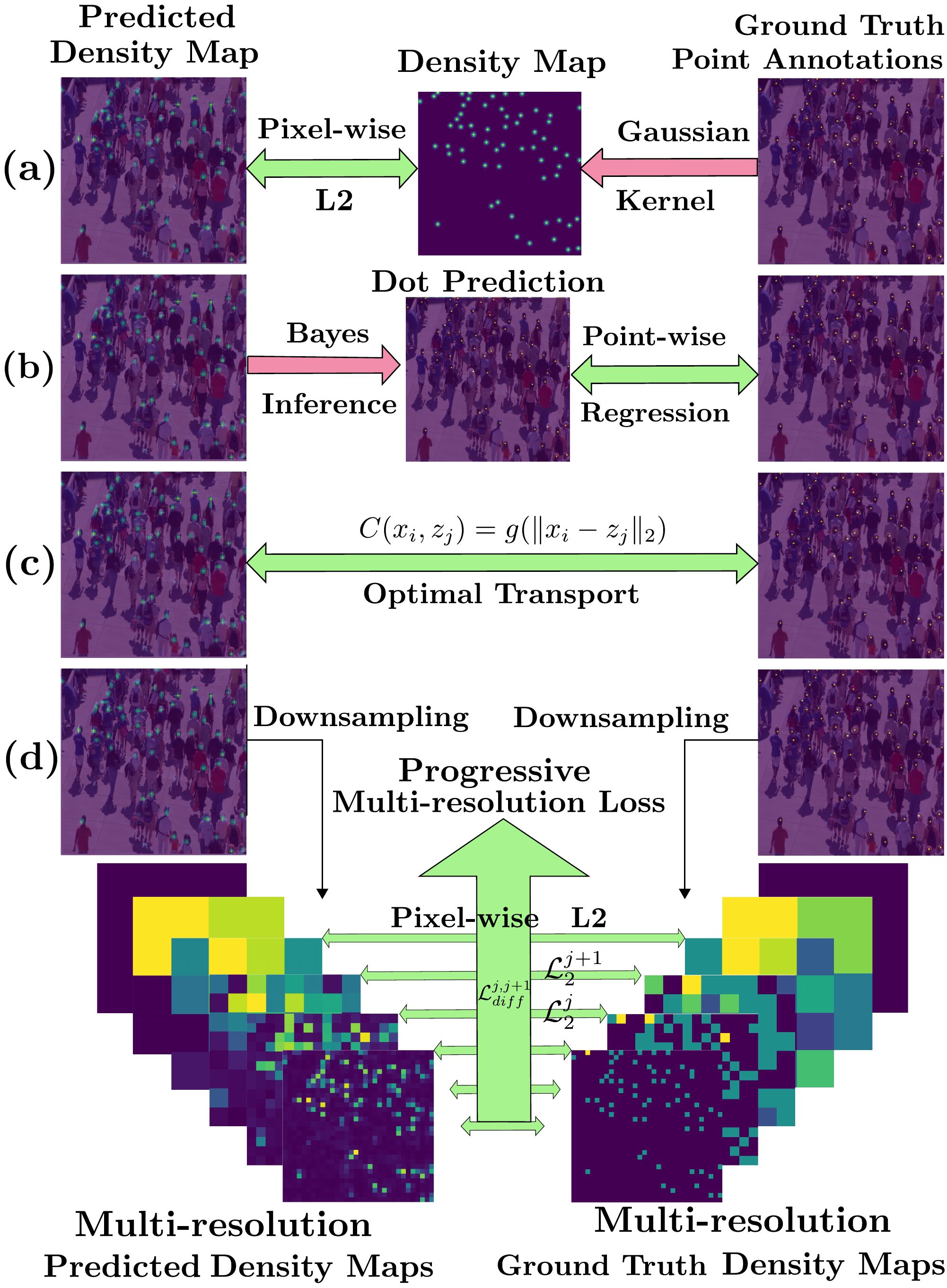}
		\caption{Modeling the correspondence of predicted density map and ground truth annotation: 
			(a) Using Gaussian kernel to convert ground truth point annotation to density map. Then pixels at the same coordinates are viewed and matched. 
			(b) Using Bayes' theorem to find the ground truth annotation for a pixel in the predicted density map.
			(c) Using Optimal Transport of a cost matrix of displacement distance to find the correspondence. 
			(d) We also view pixels at the same coordinates as matched but we propose to measure the similarity between the generated density map and the ground truth map at multiple resolutions. By maximizing its posterior probability, we derive a new loss, where the traditional single-resolution L2 loss is its particular case. 
		}
		\label{fig:compare}
	\end{figure}
	
	Currently, there are three main types of methods to establish the association of pixels between the predicted density map and the ground truth annotations: 
	(i) Gaussian kernel as shown in Fig.~\ref{fig:compare}(a). Ground truth dot annotations are converted to  a density map by placing a Gaussian kernel at each annotation point. 
	Then,  L2 distance loss  is used to narrow the gap between pixels at the same locations  of the predicted density map and the converted ground truth density map~\cite{lempitsky_learning_2010, zhang_single-image_2016, Adaptive_Density_Map, kernel-Based}.  
	These methods focus on designing algorithms for selecting the Gaussian kernel bandwidth to generate a high-quality ground truth density map.  
	(ii) Bayesian Inference (BI) as shown in Fig.~\ref{fig:compare}(b). Instead of generating density maps from point annotations, BI~\cite{ma_bayesian_2019}  computes the posterior probability of a pixel location in the predicted density map according to Bayes' theorem to infer its corresponding ground truth annotation.
	(iii) Optimal Transport (OT) as shown in Fig.~\ref{fig:compare}(c). As opposed to the above indirect approaches, a direct distance measurement based on optimal transport (OT) theory~\cite{wang_distribution_2020, wan_generalized_2021} is proposed to find the best match between the predictions and annotations in an end-to-end way. OT no longer requires the setting of Gaussian kernel bandwidth but relies on a cost matrix that reflects the distance between pixels of the predicted density map and ground truth annotation.
	Different from the above-mentioned methods, we view pixels at the same coordinates that are matched, but we propose to measure the similarity between a predicted density map and the ground truth under multiple resolutions. 
	By maximizing the posterior probability in such a setting, we derive a log-formed L2-difference based loss, where the traditional single-resolution loss is its particular case.
	We theoretically prove that the derived multi-resolution loss is superior to a single-resolution L2 loss. 
	Without bells and whistles, the proposed loss brings substantial performance improvement over baselines and favorable performance against state-of-the-art methods on four crowd counting benchmark datasets, ShanghaiTech A \& B, UCF-QNRF, and JHU-Crowd++.
	
\section{Related Work}

\noindent\textbf{Counting based on L2 regression of density map.}  
Counting based on density maps   is first proposed in~\cite{lempitsky_learning_2010}.
Unlike  previous methods of detection and localization for counting, this method transforms the counting problem into estimating object density, and the sum of the densities in a region is viewed as the total number of objects in that region. 
Density map regression is implemented by minimizing a regularized risk quadratic cost function. 
To this end, a ground truth density map is used in supervised learning. Specifically, the density map is generated by placing a fixed-bandwidth Gaussian kernel at each annotated point. However, the geometry of the scene and perspective effects can lead to distortion of the generated density map~\cite{zhang_single-image_2016}. 
When the density of objects is considered to be evenly distributed, the average distance between the center and its nearest neighbors gives a reasonable estimate of the geometric distortion. Under this assumption, ~\cite{zhang_single-image_2016} proposes to specify the bandwidth of the Gaussian kernel for each annotated point to the average distance from the neighbors.
In~\cite{Adaptive_Density_Map}, it is suggested that specific density maps 
designed manually may not be optimal for the end-to-end training process. To address this issue, a framework for generating density maps is proposed in~\cite{Adaptive_Density_Map}, which improves the quality of density maps by fusing blurred density maps generated by various Gaussian kernels. 

\noindent\textbf{Crowd counting based on Bayesian Inference.}
Bayesian loss (BL)~\cite{ma_bayesian_2019}, no longer using low-quality density maps generated by Gaussian kernels, constructs a density contribution probability model from the point annotations. 
Compared to pixel-wise L2 regression methods, L1 regression is performed point-wisely in BL between the annotated points and the contributions of the density map. 
Bayesian theory plays the role of converting prior probabilities into posterior contributions in this process. Because the computation of contributions in regions far from the annotated points is not always correct, BL introduces background modeling so that the predicted density on the background can be counted as zero. In~\cite{Congestion-Aware_Bayesian}, BL is further improved on background probability modeling, allowing BL to adapt to scale variations of people.

\noindent\textbf{Crowd counting based on Optimal Transport.}
DM-Count~\cite{wang_distribution_2020} theoretically and empirically shows that the generalization performance can be hurt by the ``ground truth'' density map smoothed by Gaussian kernels from point annotations. 
DM-Count  views crowd counting as a distribution matching problem. 
It uses Optimal Transport (OT) to find the best match between generated density map and point annotations.
The solution of the optimal transport is iteratively approximated by using the Sinkhorn algorithm~\cite{sinkhorn} with a cost matrix. 
GL~\cite{wan_generalized_2021} follows the work on matching the distribution of density maps using OT with a modified cost matrix that concerns the perspective effect in crowd images. GL changes the form of OT used in counting so that the density map matches the point annotations directly without normalization of the predicted density map, which also makes L2 and BL sub-optimal solutions to the deformed OT problem.
	
\section{Proposed Method}
In this section, we first formulate the crowd counting task as a probability maximization problem. Then we derive the optimization loss for deep learning model by maximizing the posterior probability, during which we also show the resulting loss function is superior to a single-resolution L2 loss.
\begin{figure}[tp]
    \centering
    \includegraphics[width=0.6\linewidth]{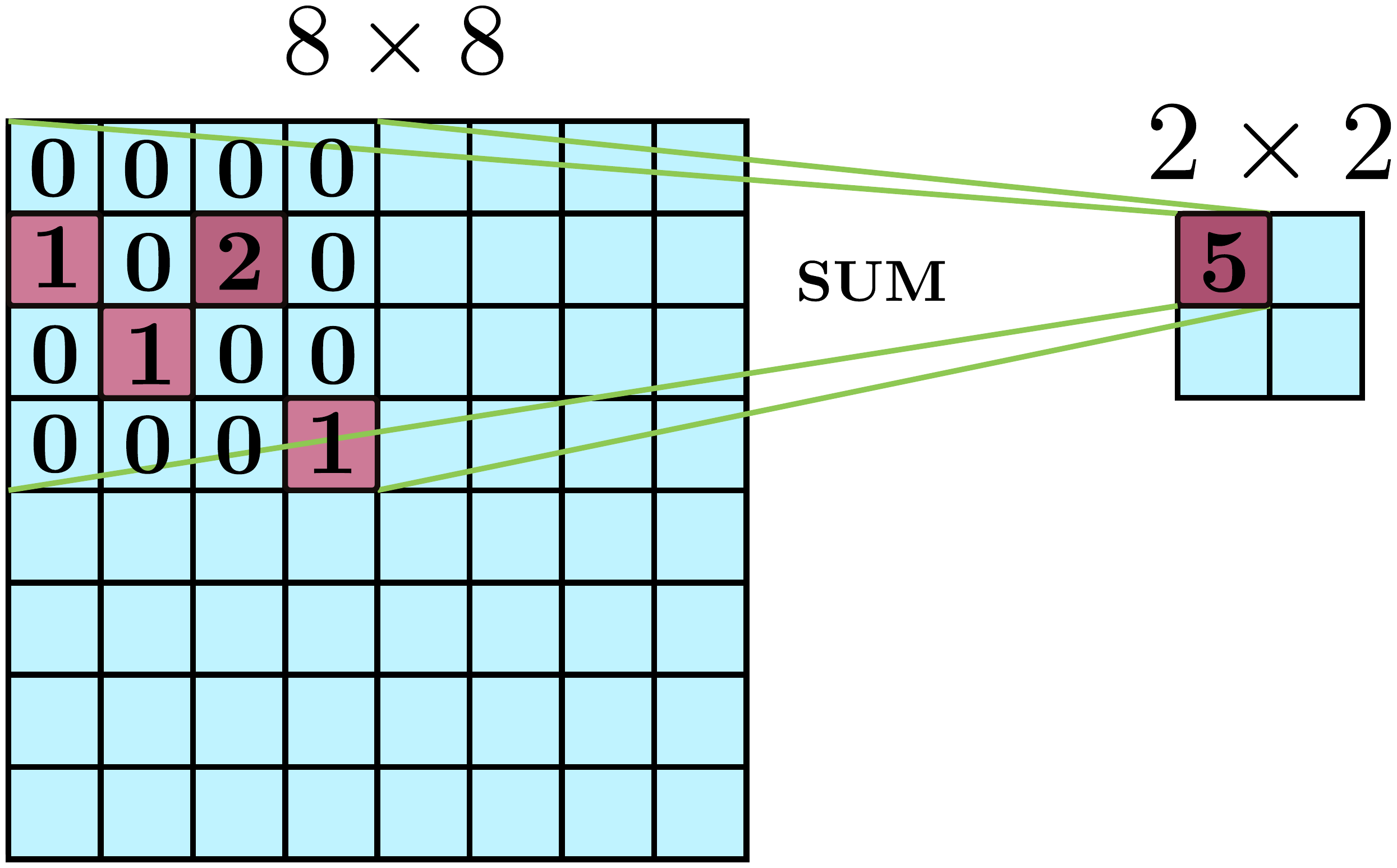}
    \caption{Downsampling density map from $ 8 \times 8$ to $2 \times 2 $.
    }
    \label{fig:downsample}
\end{figure}
    \subsection{Problem formulation}
    Let $ \hat{\boldsymbol{y}} = f(\mathbf{x}) \in\mathbb{R}^{2^L\times2^L}  $ represent the predicted density map, where $ f $ is a counting model, $ \mathbf{x}$ is an input image, and $2^L\times2^L$ is the resolution of the predicted density map.
	Let $ \boldsymbol{y}_{ori}$ be the original  point annotation map, of which each element is 1 or 0, denoting the occupancy of one person or object. We convert $ \boldsymbol{y}_{ori}$ to $\boldsymbol{y}\in\mathbb{R}^{2^L\times2^L}$ by evenly dividing it into $2^L\times2^L$ grids and the sum of the annotations in each grid is used as the new pixel value in $\boldsymbol{y}$ as shown in Fig.~\ref{fig:downsample}.
 
	In addition to measuring the similarity between $\hat{\bs{y}}$ and $\bs{y}$ on the resolution $2^L\times2^L$ as existing methods, we propose to also measure the similarity at multiple sub-resolutions: $2^i\times2^i, i \in \{n_0, n_1,\cdots, n_k\}$, where $n_i$ is an integer, $n_i<n_{i+1}$ and $k<L$.
	Predicted density map $\hat{\boldsymbol{y}}_{2^i}$ and ground truth annotation map ${\boldsymbol{y}}_{2^i}$  at  resolution $2^i\times2^i$ can be obtained by evenly dividing $\hat{\boldsymbol{y}}$ or $\boldsymbol{y}$  into $ 2^i\times 2^i $ grids and  summing the elements in each grid to get the value for each element in the new resolution. In such a setting, training a model $f$ is to maximize the posterior probability $p(\bs{y}=\bs{y}_{2^L}, \bs{y}_{2^{n_k}}, \cdots, \bs{y}_{2^{n_0}}|\mb{x})$ on resolutions $ \mathbb{N}= \{n_0, n_1,\cdots, n_k, L\}$.
     According to the chain rule of conditional probability, and because   $ \bs{y}_{2^{n_{i+1}}} $ contains all the information in $ \boldsymbol{y}_{2^{n_i}} $ and $ \boldsymbol{y} $ contains all the information in $ \boldsymbol{y}_{2^{n_k}} $, the posterior probability can be decomposed
     {\small{
	\begin{equation}
		\begin{aligned}
            p(\bs{y}|\mb{x}, \mathbb{N})
            \overset{def}{=} &
            p(\bs{y}, \bs{y}_{2^{n_k}}, \cdots, \bs{y}_{2^{n_0}}|\mb{x})\\
            =&p(\bs{y}|\bs{y}_{2^{n_k}},\mb{x})\left[\prod\limits_{j=0}^{k-1} p(\bs{y}_{2^{n_{j+1}}}|\bs{y}_{2^{n_j}},\mb{x})\right]p(\bs{y}_{2^{n_0}}|\mb{x}).
		\end{aligned}
		\label{eq:prob}
	\end{equation}
	}}
    
    The goal of training the model $f$ is equivalent to maximizing the marginal likelihood
    \begin{equation}
        f^* =  \arg\max_{f} \mathbb{E}_D \log p(\bs{y}, \bs{y}_{2^{n_k}}, \cdots, \boldsymbol{y}_{2^{n_0}}|\mb{x}).
    \end{equation}

    In the following we prove that the above marginal likelihood is further equivalent to minimizing the loss in Section~\ref{sec:loss}.
    
    \subsection{From posterior probability maximization to the loss function}
    
    In existing  methods  \cite{Bayesian_Model_Adaptation, Bayesian_Poisson, Low-Level_Features_and_Bayesian_Regression}, the posterior probability 
    of the number of objects is set to be independent, identically distributed (i.i.d.), Gaussian. Since the dimension of the $\boldsymbol{y}_{2^{n_0}}$ is $4^{n_0}=2^{n_0}\times2^{n_0}$, we additionally assumes that the pixels of $\boldsymbol{y}_{2^{n_0}}$ are 
    i.i.d. Gaussian, 
    \begin{equation}
        p(\boldsymbol{y}_{2^{n_0}}|\mb{x})\sim \mathcal{N}(\hat{\boldsymbol{y}}_{2^{n_0}},\sigma_0^2 \mb{I}_{4^{n_0}}),
        \label{eq:gaussian}
    \end{equation}
    where $\sigma_0^2$ is the parameter of Gaussian distribution, $\mb{I}_{4^{n_0}}$ is a $4^{n_0}\times 4^{n_0}$ identity matrix. When we make this assumption, the conventional single-resolution L2 loss function becomes a particular case of our posterior distribution that $ \mathbb{N}= \{n_k, L\}$ and $p(\bs{y}|\boldsymbol{y}_{2^{n_k}},\mb{x}) =  p(\bs{y}|\boldsymbol{y}_{2^{n_k}}) $. 

    We aim to learn the high-resolution density maps without destroying the lower-resolution maps as much as possible. In order to keep the low-density map unaffected, We define that the residual density map  $\boldsymbol{r}_{{j_1}, {j_2}}$, $\hat{\boldsymbol{r}}_{{j_1}, {j_2}}$ between two density maps with different resolutions $2^{j_1}$ and $2^{j_2}$ , 
    \begin{align}
        \boldsymbol{r}_{{j_1}, {j_2}} & \overset{def}{=} \boldsymbol{y}_{2^{{j_2}}} - 4^{{j_1}-{j_2}}\mathbf{U}(\boldsymbol{y}_{2^{{j_1}}}, 2^{j_2-j_1}) ,\\
        \hat{\boldsymbol{r}}_{{j_1}, {j_2}} & \overset{def}{=} \hat{\boldsymbol{y}}_{2^{{j_2}}} - 4^{{j_1}-{j_2}}\mathbf{U}(\hat{\boldsymbol{y}}_{2^{j_1}}, 2^{j_2-j_1}),
    \end{align}
    where $j_1<j_2$, $\mb{U}(\cdot, 2^{j_2-j_1})$ is the replication upsampling function of scale $2^{j_2-j_1}$.
    We assume that the residual maps $\boldsymbol{r}_{n_{j}, n_{j+1}}$  are generated from Gaussian distribution, which means
    \begin{equation}
        \begin{aligned}
             p(\boldsymbol{r}_{n_{j}, n_{j+1}} |\mb{x}) 
            \sim  \mathcal{N}(\hat{\boldsymbol{r}}_{n_{j}, n_{j+1}} , \sigma_{j+1}^2\mb{I}_{4^{n_{j+1}}}),
        \end{aligned}
        \label{fen}
    \end{equation}
    where $ \sigma_{j+1}^2 $ is the variance parameter of the Gaussian distributions. $ \bs{r}_{n_{j+1}, n_j}  $ needs to satisfy the priors that $ \mb{D}(r_{n_{j+1}, n_j}, 2^{n_{j+1}- n_j}) = \mb{0}_{4^{n_j}}$, where $\mb{0}_{4^{n_j}}$ is the zero vector of length $4^{n_j}$, $\mb{D}(\cdot, 2^{n_{j+1}- n_j})$ is the average downsampling function of scale $2^{n_{j+1}-n_j}$. The conditional probability
    $p(\boldsymbol{y}_{2^{n_{j+1}}} | \boldsymbol{y}_{2^{n_j}} , \mb{x})$ has the following form,
    \begin{equation}
    \begin{aligned}
        &p(\boldsymbol{y}_{2^{n_{j+1}}} | \boldsymbol{y}_{2^{n_j}} , \mb{x}) 
        =  p(\boldsymbol{r}_{{n_{j}}, n_{j+1}} | \boldsymbol{y}_{2^{n_j}} , \mb{x})\\
        =& \frac{
            p(\boldsymbol{r}_{{n_{j}}, n_{j+1}} |, \mb{x})
        }{
        \int p(\boldsymbol{v}_{{n_{j+1}}, n_j} |, \mb{x}) \mathbb{I}(\mb{D} (\boldsymbol{v}_{{n_{j+1}}, n_j}) = \mb{0}_{4^{n_{j}}}) d \boldsymbol{v}
        }\\
        =& \frac{
            \mathcal{N}(\hat{\boldsymbol{r}}_{n_{j}, n_{j+1}} , \sigma_{j+1}^2\mb{I}_{4^{n_{j+1}}})
        }{
        \mathcal{N}(\boldsymbol{0}_{4^{n_j}} , 
        4^{n_{j+1}-n_j} \sigma_{j+1}^2\mb{I}_{4^{n_{j}}})
        }\\
        =&  
        \sqrt{\frac{4^{(n_{j}-n_{j+1})4^{n_j}}}{
        (2\pi \sigma_{j+1}^2)^{4^{n_{j+1}-n_j}} 
        }}
        \exp\{ -\frac{1}{2\sigma_{j+1}^2} 
        \| \hat{\boldsymbol{r}}_{n_{j}, n_{j+1}} - \boldsymbol{r}_{n_{j}, n_{j+1}} \|_2^2
        \},
    \end{aligned}
    \label{eq:conditional}
    \end{equation}
    where $ \mathbb{I}(\cdot) $ is the indicator function.

    Combining Eq.~(\ref{eq:prob})(\ref{eq:gaussian})(\ref{eq:conditional}), the marginal likelihood of the training data $D$ equals to a linear combination of $\mathcal{L}_{diff}^{n_{j}, n_{j+1}}, j=0, \cdots, k-1$ ,and $\mathcal{L}_2^{n_0}$:
    \begin{equation}
        \begin{aligned}
            \mathbb{E}_D & \log p(\boldsymbol{y}|\mathbf{x}, \mathbb{N}) =  \mathbb{E}_D \  \log p(\boldsymbol{y}|\boldsymbol{y}_{2^{n_k}}, \mb{x})-\\
            &\dfrac{1}{2}\sum_{j=1}^{k-1}[
                \frac{1}{\sigma_{j+1}^2} 
                \mathcal{L}_{diff}^{n_{j}, n_{j+1}} 
                + (4^{n_{j+1}} - 4^{n_j}) \log(2\pi \sigma_{j+1}^2) 
                \\
                +& (n_{j+1}-n_j) 4^{n_j}\log 4
            ]
                - \frac{1}{2}(
                    \frac{1}{\sigma_0^2}
                    \mathcal{L}_2^{n_0}
                    + 4^{n_{0}} \log(2\pi \sigma_{0}^2)
                ),
        \end{aligned}
        \label{6}
    \end{equation}
    where 
    \begin{align}
        &\mathcal{L}^i_2 \overset{def}{=} \mathbb{E}_D  \|\hat{\boldsymbol{y}}_{2^i} - \boldsymbol{y}_{2^i}\|_2^2,\\
            &\begin{aligned}
                \mathcal{L}^{n_j,n_{j+1}}_{diff} \overset{def}{=}&	 
             \mathcal{L}^{n_{j+1}}_2 - 4^{n_{j}-n_{j+1}}\mathcal{L}^{n_j}_2
                \\=&
                \mathbb{E}_D
                 \|
         \boldsymbol{r}_{n_{j}, n_{j+1}} - \hat{\boldsymbol{r}}_{n_{j}, n_{j+1}}	
            \|_2^2.
            \end{aligned}
            \label{eq:diff}
    \end{align}

    \textbf{Remark}: Only the density map-based methods are discussed in this paper, so the model does not include a localization branch. Therefore, $p(\boldsymbol{y}|\boldsymbol{y}_{2^{n_k}}, \mb{x})$ does not appear in the final loss function. In the case of 
    crowd localization, $\mathbb{E}_D \  \log p(\boldsymbol{y}|\boldsymbol{y}_{2^{n_k}}, \mb{x})$ corresponds to the loss function of the localization branch.

    \subsection{Selection of $\mathbb{N}$}
    \label{choice-n}
    
    By calculating the derivatives of Eq.~(\ref{6}), the optimal solution of $ \boldsymbol{\sigma} = \{\sigma_{n_j}|j=0,\cdots,k\} $ is obtained,
	\begin{equation}
            \begin{aligned}
                \sigma_{j}^2 & = \frac{\mathcal{L}_{diff}^{n_{j-1}, n_{j}}}{4^{n_j}-4^{n_{j-1}}}, \forall j \in \{1, \cdots, k\},\\
                \sigma_{0}^2 & = 4^{-n_0} \mathcal{L}_2^{n_0}.
            \label{8}
            \end{aligned}
        \end{equation}
	Then taking it to Eq.~(\ref{6}), the marginal likelihood of the training data can be simplified,
	\begin{equation}
		\begin{aligned}
			\max\limits_{\boldsymbol{\sigma}}\mathbb{E}_D & \log p(\boldsymbol{y}|\mathbf{x}, \mathbb{N}) \\
                = &\mathbb{E}_D \  \log p(\boldsymbol{y}|\boldsymbol{y}_{2^{n_k}},\mb{x})
			-\dfrac{-1+2\pi}{2} 4^{n_k} \\
			&
			- \dfrac{1}{2}\sum_{j=1}^{k}(4^{n_j}-4^{n_{j-1}})
			\log \dfrac{4^{n_j} \mathcal{L}_{diff}^{n_{j-1}, n_{j}}}{4^{n_j}-4^{n_{j-1}}}\\
			&-\dfrac{1}{2} 4^{n_0} \log \mathcal{L}_2^{n_0}
			,
		\end{aligned}
		\label{9}
	\end{equation}
	where $ \boldsymbol{\sigma}  = \{\sigma_0, \cdots, \sigma_{k}\} $. 
	To guide the choice of $\mathbb{ N} $, we prove the following theorem.
	\begin{theorem}
            \label{thm:N}
		Let  
		$ \mathbb{N}'=\{0, 1, \cdots, n_k\} \cup \{L\}$. Then the log-likelihood function in case of $ \mathbb{N}' $ is greater than that of $ \mathbb{N}$,
		\begin{equation}
			\max\limits_{\boldsymbol{\sigma}}\mathbb{E}_D \log p(\boldsymbol{y}|\mathbf{x},\mathbb{N}) 
            \leq \max\limits_{\boldsymbol{\sigma}}\mathbb{E}_D \log p(\boldsymbol{y}|\mathbf{x},\mathbb{N}').
			\label{10}
		\end{equation}
		
	\end{theorem}
    The proof of this theorem can be found in the supplementary material.
        
    \ul{This theorem tells us that measuring the predicted density map at $n_k$ additional resolutions starting from $2^0\times2^0$ leads to a better fitting of the training data than merely using resolution $2^{n_k}\times2^{n_k}$.}
	
	When $ \mathbb{N} =\{0, 1, \cdots, n \} \cup \{L\}$, the Eq.~(\ref{9}) is equal to 
        \begin{equation}
		\begin{aligned}
			&\max\limits_{\boldsymbol{\sigma}}\mathbb{E}_D \log p(\boldsymbol{y}|\mathbf{x}, \mathbb{N})
			=  \mathbb{E}_D \  \log p(\boldsymbol{y}|\boldsymbol{y}_{2^{n}}, \mb{x})
			-\dfrac{-1+2\pi}{2} 4^{n} \\
			&- \dfrac{1}{2}\sum_{j=1}^{n}(4^{j}-4^{{j-1}})
			\log \dfrac{4}{3} \mathcal{L}_{diff}^{{j-1}, {j}}
                -\dfrac{1}{2} \log \mathcal{L}_2^{0}
			.
		\end{aligned}
		\label{loss}
	\end{equation}
    \begin{figure}[tp]
        \centering
        \includegraphics[width=.8\linewidth]{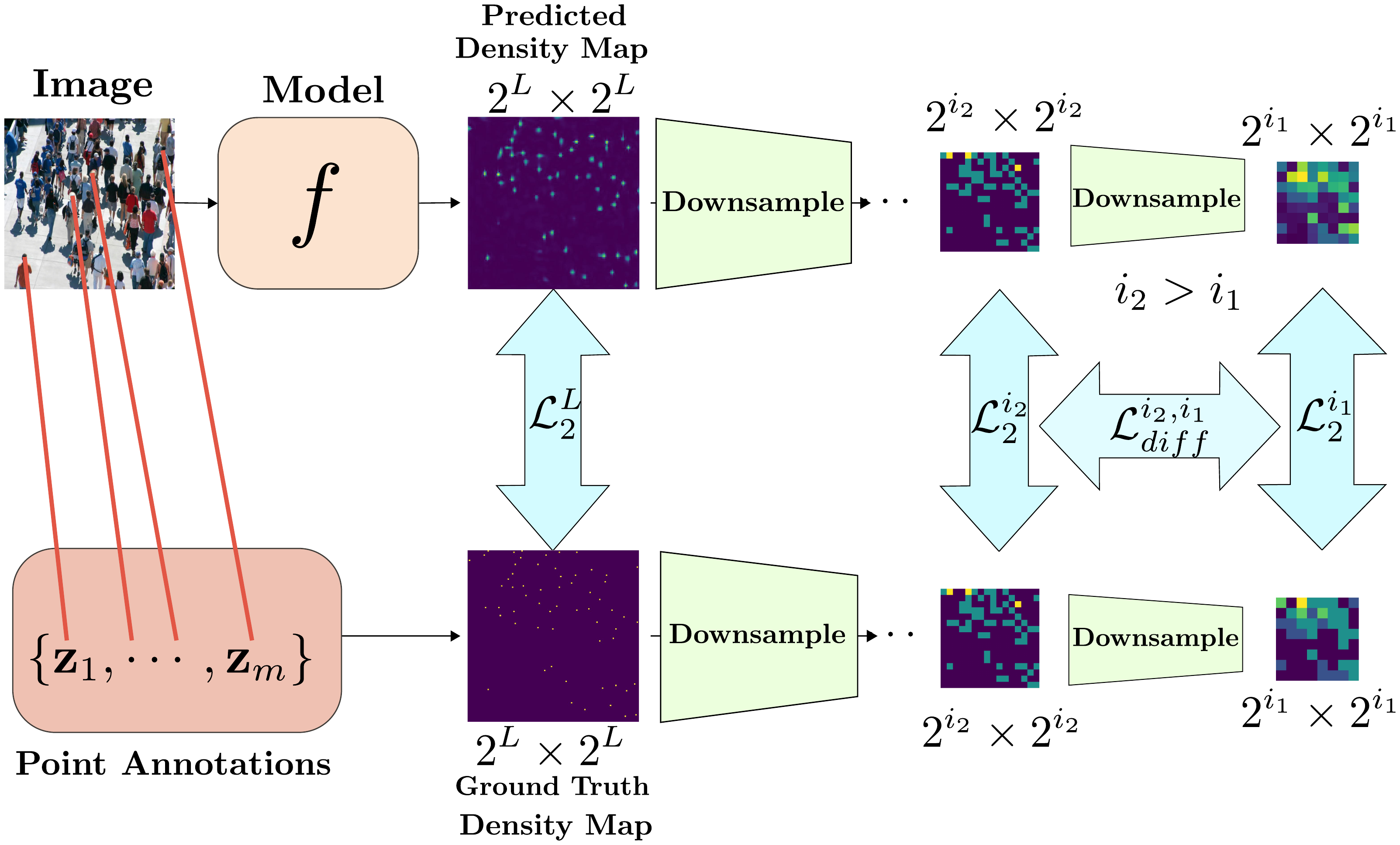}
        \caption{Illustration of  multi-resolution loss.
        }
        \label{fig:loss}
    \end{figure}
    \subsection{Regularization}
    In the previous analysis, no additional restrictions are added to the variance $\sigma_j^2$. Therefore, we add penalty terms to regularize them to limit the range of variance,
    \begin{equation}
        \begin{aligned}
            \mathcal{L}_{reg} = \sum_{j=0}^n \beta_j \sigma_j^2 .
        \end{aligned}
        \label{loss-reg-ori}
    \end{equation}
    Since this form is too complex and has too many parameters, we simplify it to one regular term $\mathcal{L}_2^{n} $ by taking appropriate $\beta_0, \cdots, \beta_n$. 
       
    In Eq.~(\ref{loss}), the term $ \log \mathcal{L}_{diff}^{n, n-1} $ has the largest coefficient, which makes model learning mainly based on this resolution. Therefore, We want the model to learn by combining multiple resolutions. 
    
    Because the second two terms in Eq.~(\ref{loss}) are constants, our loss function is obtained by re-weighting the coefficients of the rest of the equation. The process of re-weighting coefficients is equivalent to a weighted average of log-likelihood functions with different $n$. The formulaic representation is as follows,
    \begin{equation}
    \begin{aligned}
         &\sum_{k=0}^n \alpha_k [ \sum_{j=1}^{k}(4^{j}-4^{j-1})
        \log \mathcal{L}_{diff}^{j-1, j} + \log \mathcal{L}_2^0]   \\
        = &  \log \mathcal{L}_2^0 +  \sum_{j=1}^{n} 
        \log \mathcal{L}_{diff}^{j-1, j} ,
    \end{aligned}
    \end{equation}
    where $ \alpha_k$ is the coefficient for the resolution of $2^k\times 2^k$ and $\alpha_0 ,\cdots, \alpha_n$ are the solution of $  (4^j - 4^{j-1})\sum_{k=j}^n \alpha_k = 1, \forall j \in \{1, \cdots, n\}$, $ \sum_{k=0}^n \alpha_k = 1 $.

    \subsection{Progressive Multi-resolution Loss}
    \label{sec:loss}
    
    The progressive multi-resolution loss is defined as
	\begin{equation}
		\mathcal{L}_{PML} = \log \mathcal{L}^0_2+ \sum_{j=1}^{n} \log \mathcal{L}^{j-1, j}_{diff}.
		\label{loss-pml}
	\end{equation}
 
    When the regular term $\mathcal{L}_2^L$ is added, the final loss function is defined by:
    	 \begin{equation}
    	 	\mathcal{L} = \mathcal{L}_{PML} + \mathcal{L}_2^L.
                \label{loss-final}
    	 \end{equation}
    Fig.~\ref{fig:loss} shows an illustration of the loss   used in the proposed model.
	
  \begin{table*}[!htbp]
    \centering
     
    \rowcolors{1}{}{lightgray}
    \resizebox{\linewidth}{!}{
    \begin{tabular}{l ll|ll|ll|ll}
         \toprule
        
        & \multicolumn{2}{c}{JHU-Crowd++} \vline
        & \multicolumn{2}{c}{UCF-QNRF} \vline
        & \multicolumn{2}{c}{ShanghaiTech A} \vline
        & \multicolumn{2}{c}{ShanghaiTech B}\\
        & MAE  $\downarrow$ & MSE  $\downarrow$
        & MAE  $\downarrow$ & MSE  $\downarrow$
        & MAE  $\downarrow$ & MSE  $\downarrow$
        & MAE  $\downarrow$ & MSE  $\downarrow$ \\
        \midrule
        FIDT ~\cite{FIDT} 
        &66.6 &253.6
        & 89.0 & 153.5
        & 57.0 &103.4
        & 6.9 &11.8
        \\
        Ours + FIDT
        & 62.3(\textcolor{red}{-4.3}) & 261.9(+8.3)
        & 87.2(\textcolor{red}{-1.8}) & 151.9(\textcolor{red}{-1.6})
        &54.5(\textcolor{red}{-2.5}) & 92.7(\textcolor{red}{-10.7})
        &6.9(-0.0) &  9.8(\textcolor{red}{-2.0})
        \\
        FDC-18 ~\cite{MFDC} 
        &- &-
        &93.0 &157.3
        &65.4 & 109.2
        & 11.4 & 19.1
        \\
        Ours + FDC-18
        & - & -
        &82.1(\textcolor{red}{-10.9}) & 153.5(\textcolor{red}{-3.8})
        &62.3(\textcolor{red}{-3.1}) & 101.1(\textcolor{red}{-8.1})
        &6.7(\textcolor{red}{-4.7}) &  10.8(\textcolor{red}{-8.3})
        \\
        FDC-ConvNeXtS
        & 61.3 & 275.2
        & 83.2 & 156.7
        & 59.2 & 97.3
        & 7.0 &  11.1    
        \\
        Ours + FDC-ConvNeXtS
        & 55.2(\textcolor{red}{-6.1}) & 232.4(\textcolor{red}{-42.8})
        & 79.9(\textcolor{red}{-3.3}) & 132.5(\textcolor{red}{-24.2})
        & 53.4(\textcolor{red}{-5.8})  & 93.1(\textcolor{red}{-4.2})
        & 6.2(\textcolor{red}{-0.8})  & 9.7(\textcolor{red}{-1.4})
        \\
        \bottomrule
    \end{tabular}
    }
    \caption{Combination with open source state-of-the-art methods. \textcolor{red}{RED} indicates the decrease in MAE and MSE compared with the original methods. }
    \label{tab:combine-with-model}
\end{table*}


\section{Experiments}
In this section, we present experiments using our loss function, which contains comparisons with SOTA methods and ablation studies.
    \subsection{Experiment settings}
        
    \begin{table}[!htbp]
    \centering
     \rowcolors{1}{}{lightgray}
    \begin{tabular}{l c c c}
        \toprule
        Datasets & Number of objects & \#Train & \#Test\\
        \midrule
        JHU-Crowd++~\cite{JHU} & [0, 7,286] &2,722&1,600\\
        UCF-QNRF~\cite{UCF-QNRF} & [49, 12,865] & 1,201 & 334\\
        ShanghaiTech A~\cite{zhang_single-image_2016} & [33, 3,139] & 482 & 300\\
        ShanghaiTech B~\cite{zhang_single-image_2016} & [9, 578] & 716& 400\\
        \bottomrule
    \end{tabular}
    \caption{Crowd counting benchmarks. The second column represents the range of the number of objects per image. \#Train and \#Test represent the size of the training and test image sets. }
        \label{data}
    \end{table}
    
    \begin{table*}[!htbp]
    \centering
    \rowcolors{1}{}{lightgray}
    \begin{tabular}{lccc|cc|cc|cc }
        \toprule
        &
        & \multicolumn{2}{c}{JHU-Crowd++} \vline
        & \multicolumn{2}{c}{UCF-QNRF} \vline
        & \multicolumn{2}{c}{ShanghaiTech A} \vline
        & \multicolumn{2}{c}{ShanghaiTech B}\\
        &
        & MAE  $\downarrow$ & MSE  $\downarrow$
        & MAE  $\downarrow$ & MSE  $\downarrow$
        & MAE  $\downarrow$ & MSE  $\downarrow$
        & MAE  $\downarrow$ & MSE  $\downarrow$\\
        \midrule
        BL ~\cite{ma_bayesian_2019}
        & \scriptsize ICCV’19 
        &75.0 &299.9 
        &88.7 &154.8 
        &62.8 &101.8 
        &7.7 &12.7
        \\
        NoiseCC ~\cite{wan_modeling_2020}
        & \scriptsize NeurIPS’20 
        &67.7 &\underline{258.5 }
        &85.8 &150.6 
        &61.9 &99.6 
        &7.4 &\underline{11.3}
        \\
        DM count ~\cite{wang_distribution_2020}
        & \scriptsize NeurIPS’20 
        &68.4 &283.3 
        &85.6 &148.3 
        &\textbf{59.7} &\underline{95.7}
        &7.4 &11.8
        \\
        GL ~\cite{wan_generalized_2021}
        & \scriptsize CVPR’21 
        &\underline{59.9} &259.5 
        &\underline{84.3} &\underline{147.5 }
        &61.3 &\textbf{95.4} 
        &\underline{7.3} &11.7
        \\
        Ours
        &
        &\textbf{57.5} &\textbf{227.0}
        & \textbf{80.1} & \textbf{131.2} 
        & \underline{61.1} &104.8
        &\textbf{6.6} &\textbf{10.8}
        \\
        \bottomrule
    \end{tabular}
    \caption{Comparison with state-of-the-art loss functions.}
    \label{tab:sota-loss}
    \end{table*}
    
    \begin{figure*}[!htbp]
    \centering
    \includegraphics[width=1\linewidth]{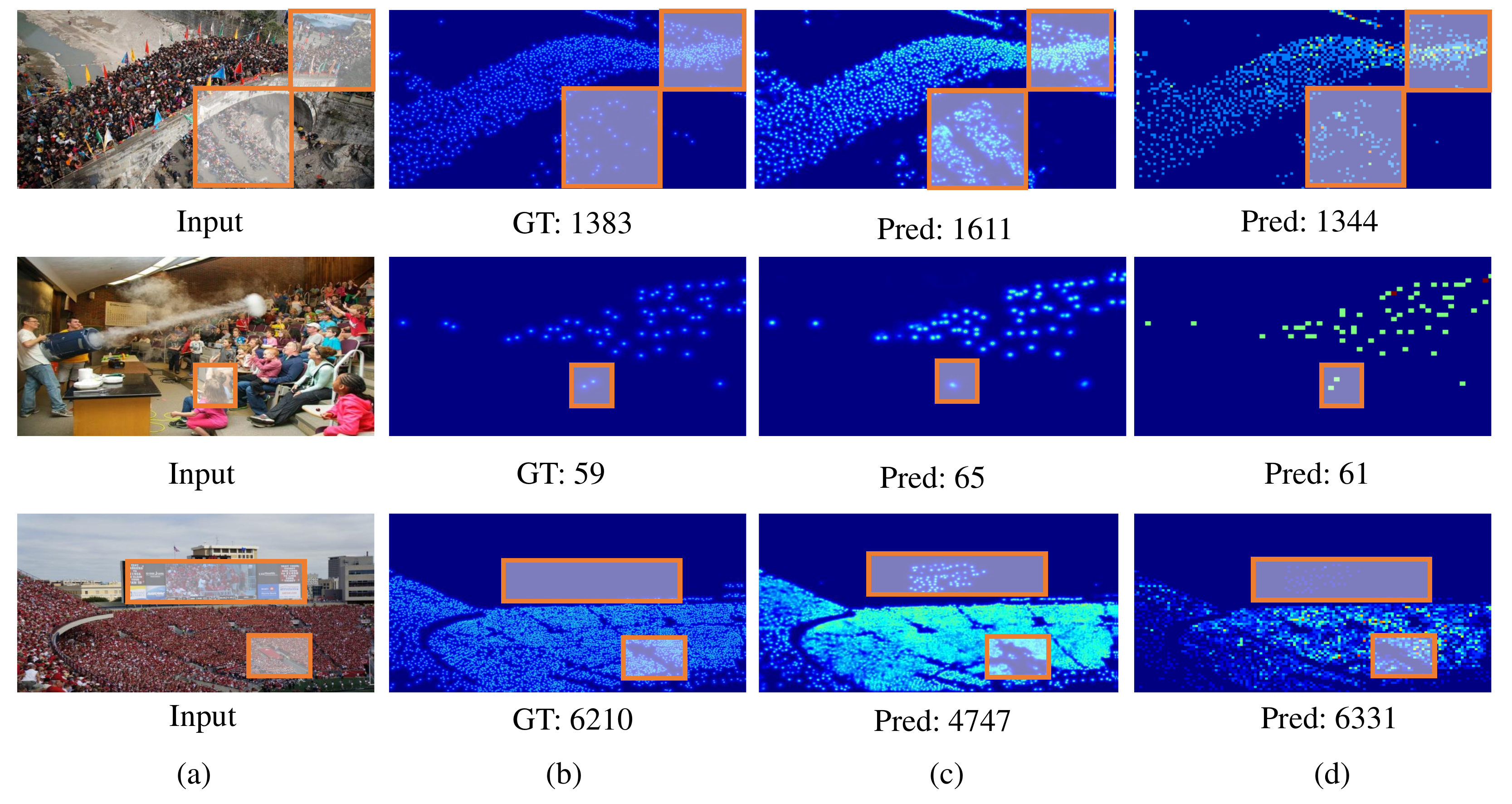}
    \caption{
            Density maps of our loss function combined with FIDT~\cite{FIDT} on JHU-Crowd++. Differences in density maps are highlighted with rectangular areas.
            (a) Input images. (b) The ground truth density maps. (c) The predicted density maps of FIDT. (d) The predicted density maps of our loss function combined with FIDT. 
    }
    \label{fig:quant}
    \end{figure*}
    
\noindent \textbf{Dataset and Augmentation.} 
	Our proposed loss function is evaluated on four datasets of crowd people counting: ShanghaiTech A and B ~\cite{zhang_single-image_2016}, UCF-QNRF ~\cite{UCF-QNRF}, and JHU-Crowd++ ~\cite{JHU}. Except for JHU-Crowd++, which has a size of 500 verification sets, the other three datasets have no verification sets. Table~\ref{data} summarizes the datasets.
	During the training, we sequentially use the following data augmentation: Color Jitter, Rand Scaling (from 0.9 to 1.1), Random crop (to 1024x1024), Random Rotate (from 0 to 45), and Horizontal Flip (with the probability of 0.5). In the inference phase, we limit the longest edge within 1536 and pad the image size to an integer multiple of 64.
~\\

\noindent  \textbf{Criteria.} The Mean Absolute Error (MAE) and the Root Mean Square Error (MSE) are measures of the performance of the models in previous works ~\cite{mse1, SFCN}:
\begin{equation}
    \begin{aligned}
        MAE &= \frac{1}{K} \sum_{k=1}^K |\hat{m}_k-m_k|, \\
        MSE &= \sqrt{\frac{1}{K}\sum_{k=1}^K (\hat{m}_k-m_k)^2 },
    \end{aligned}
\end{equation}
where $\hat{m}_k$ and $ m_k$ are the estimated counting number and ground truth counting number, the size of the entire test dataset is $K$. 
~\\

\noindent \textbf{Network Structures.} 
\label{network}
We use VGG16~\cite{vgg}, CSRNet~\cite{csrnet}, ResNet18~\cite{resnet}, ConvNeXtS~\cite{ConvNeXt}, and HRNet-W48~\cite{HRNet} as the backbone of our models. 
VGG16 is a standard pre-trained model for crowd counting, which uses only 3×3 convolution. We use VGG16 to extract the features and then use two 1×1 convolutional layers with ReLU activation functions as the prediction header.  
CSRNet, designed for crowd counting with VGG16 as the backbone, employs four dilated convolution layers with 512, 256, 128, and 64 channels in the prediction head. 
ResNet18 adds direct connection channels to the network, allowing information to skip a layer directly.
ConvNeXt is a pure convolution model that can compete with vision Transformers~\cite{vit, Swin} across multiple computer vision benchmarks. ConvNeXtS is a variant of ConvNeXt with 96, 192, 384, and 768 channels in the backbone. 
We add the same counting head in CSRNet to ResNet18 and ConvNext to estimate the density maps. 
HRNet is a backbone widely used in localization task . HRNet-W48 is its pre-trained model on ImageNet~\cite{imagenet}, and we use it to provide multi-scale density map output. 
~\\
	
\noindent \textbf{Hyper-parameters.} The Adam optimizer updates parameters with an initial learning rate $ 10^{-4} $. 
As our loss gradient increases with the training epochs, we adopt gradient clipping of 10 in the back-propagation.
$n$ in Eq.~(\ref{loss-pml}) is set to 4, which means we use the density maps with resolutions of $ 1\times 1 $, $ 2\times 2 $, $ 4\times 4 $, $ 8\times 8 $, and $ 16\times 16 $ in the loss function.  

    \subsection{Combination with state-of-the-art methods}

    We combine our loss function with open source SOTA methods (FIDT~\cite{FIDT} and FDC~\cite{MFDC}) by using our loss to replace L2 loss in the counting regression of these methods. 
    For FDC, we reproduce the experiment with ConvNeXtS as the backbone network and give a better result than the original method with ResNet18~\cite{resnet}. 
    Table~\ref{tab:combine-with-model} shows the fluctuations in MAE and MSE contributed from our loss function, where MAEs decrease across all data sets and methods.  
    
     Fig.~\ref{fig:quant} shows the density maps of FIDT combined with our method. The density maps of the original FIDT method deviate severely from the ground truth density maps in some regions.
    Because our loss concerns the multi-resolution, no significant deviations appear in these regions.

    \begin{table*}[tp]
    \centering
    \resizebox{\linewidth}{!}{
    \rowcolors{1}{}{lightgray}
    \begin{tabular}{lcc|cc|cc|cc|cc }
        \toprule
        &
        & 
        & \multicolumn{2}{c}{JHU-Crowd++} \vline
        & \multicolumn{2}{c}{UCF-QNRF} \vline
        & \multicolumn{2}{c}{ShanghaiTech A} \vline
        & \multicolumn{2}{c}{ShanghaiTech B}\\
        &
        & \multirow{-2}*{Backbone}
        & MAE  $\downarrow$  & MSE  $\downarrow$
        & MAE  $\downarrow$ & MSE  $\downarrow$
        & MAE  $\downarrow$ & MSE  $\downarrow$
        & MAE  $\downarrow$ & MSE  $\downarrow$\\
        \midrule
        MCNN ~\cite{zhang_single-image_2016}
        & \scriptsize CVPR’16 
        & -
        &188.9 &483.4 
        &277.0 &426.0 
        &110.2 &173.2 
        &26.4 &41.3
        \\
        SwitchCNN ~\cite{switchcnn}
        & \scriptsize CVPR’17
        & -
        &- &- 
        &228.0 &445.0 
        &90.4 &135.0 
        &21.6 &33.4
        \\
        CSRNet  ~\cite{csrnet}
        & \scriptsize CVPR’18 
        & VGG16
        &85.9 &309.2 
        &110.6 &190.1 
        &68.2 &115.0 
        &10.6 &16.0
        \\
        SANet ~\cite{ferrari_scale_2018}
        & \scriptsize  ECCV’18 
        & -
        &91.1 &320.4 
        &- &- 
        &67.0 &104.5 
        &8.4 &13.6
        \\
        CAN ~\cite{CAN}
        & \scriptsize CVPR’19 
        & VGG16
        &100.1 &314.0 
        &107 &183 
        &62.3 &100.0 
        &7.8 &12.2
        \\
        SFCN ~\cite{SFCN}
        & \scriptsize CVPR’19 
        & ResNet101
        &77.5 &297.6 
        &102.0 &171.4 
        &64.8 &107.5 
        &7.6 &13.0
        \\
        MBTTBF ~\cite{MBTTBF}
        & \scriptsize ICCV’19 
        & VGG16
        &81.8 &299.1 
        &97.5 &165.2 
        &60.2 &94.1 
        &8.0 &15.5
        \\
        BL ~\cite{ma_bayesian_2019}
        & \scriptsize ICCV’19 
        & VGG19
        &75.0 &299.9 
        &88.7 &154.8 
        &62.8 &101.8 
        &7.7 &12.7
        \\
        KDMG ~\cite{KDMG}
        & \scriptsize TPAMI’20 
        & -
        &69.7 &268.3 
        &99.5 &173.0 
        &63.8 &99.2 
        &7.8 &12.7
        \\
        LSC-CNN ~\cite{LSCCNN}
        & \scriptsize TPAMI’21 
        & -
        &112.7 &454.4 
        &120.5 &218.2 
        &66.5 &101.8 
        &7.7 &12.7
        \\
        RPNet ~\cite{rpnet}
        & \scriptsize CVPR’20 
        &  ResNet-101
        &- &- 
        &- &- 
        &61.2 &96.9 
        &8.1 &11.6
        \\
        AMRNet ~\cite{AMRNet}
        & \scriptsize ECCV’20 
        & VGG16
        &- &- 
        &86.6 &152.2 
        &61.6 &98.4 
        &7.0 &11.0
        \\
        NoiseCC ~\cite{wan_modeling_2020}
        & \scriptsize NeurIPS’20 
        & VGG19
        &67.7 &258.5 
        &85.8 &150.6 
        &61.9 &99.6 
        &7.4 &11.3
        \\
        DM count ~\cite{wang_distribution_2020}
        & \scriptsize NeurIPS’20 
        & VGG19
        &68.4 &283.3 
        &85.6 &148.3 
        &59.7 &95.7 
        &7.4 &11.8
        \\
        LA-Batch ~\cite{LA-batch}
        & \scriptsize TPAMI’21 
        & VGG16
        &- &- 
        &113.0 &210.0 
        &65.8 &103.6 
        &8.6 &14.0
        \\
        AutoScale ~\cite{Autoscale}
        & \scriptsize IJCV’21 
        & VGG16
        &65.9 &264.8 
        &104.4 &174.2 
        &65.8 &112.1 
        &8.6 &13.9
        \\
        GL ~\cite{wan_generalized_2021}
        & \scriptsize CVPR’21 
        & VGG19
        &59.9 &259.5 
        &84.3 &147.5 
        &61.3 &95.4 
        &7.3 &11.7
        \\
        P2PNet ~\cite{P2PNet}
        & \scriptsize ICCV’21 
        & VGG16
        &- &- 
        &85.3 &154.5 
        &\textbf{52.7}  &\textbf{85.1}  
        &6.2 &9.9
        \\
        SDA+BL ~\cite{SDA}
        & \scriptsize ICCV’21 
        & VGG19
        &62.6 &264.1 
        &83.3 &143.1 
        &58.4 &95.7 
        &- &-
        \\
        FIDT ~\cite{FIDT} & \scriptsize arxiv
        & HRNet-W48
        &66.6 &253.6
        & 89.0 & 153.5
        & 57.0 &103.4
        & 6.9 &11.8
        \\
        FDC-18 ~\cite{MFDC} 
        & \scriptsize ICCV’21 
        & ResNet-18
        &- &-
        &93.0 &157.3
        &65.4 & 109.2
        & 11.4 & 19.1
        \\
        FDC-ConvNeXtS &
        & ConvNeXtS
        & 61.3 & 275.2
        & 83.2 & 156.7
        & 59.2 & 97.3
        & 7.0 &  11.1    
        \\
        Ours + FIDT &
        & HRNet-W48
        & 62.3 & 261.9
        & 87.2 & 151.9
        &54.5 & \underline{92.7}
        &6.9 &  \underline{9.8}
        \\
        Ours + FDC-18 &
        & ResNet-18
        & - & -
        &82.1 & 153.5
        &62.3 & 101.1
        &6.7 &  10.8
        \\
        Ours + FDC-ConvNeXtS &
        & ConvNeXtS
        & \textbf{55.2} & \underline{232.4}
        & \textbf{79.9} & \underline{132.5}
        & \underline{53.4}  & 93.1
        & \textbf{6.2}  & \textbf{9.7} 
        \\
        Ours + VGG16
        &
        &VGG16
        & \underline{57.5} &\textbf{227.0}
        & \underline{80.1} & \textbf{131.2} 
        &61.1 &104.8
        &6.6 &10.8
        \\
        \bottomrule
    \end{tabular}
    }
    \caption{Comparison with state-of-the-art crowd Convnet methods.}
        \vspace{2mm}
    \label{tab:all-methods}
\end{table*}

    \subsection{Comparison with state-of-the-art loss functions}
    To evaluate the counting performance of our loss function, we compare VGG16 mentioned in Sec.~\ref{network} , which is trained by our loss defined in Eq.~(\ref{loss-final}) with previous state-of-art loss functions. Quantitative results on JHU-Crowd++, UCF-QNRF, ShanghaiTech A, and ShanghaiTech B are shown in Table~\ref{tab:sota-loss}. We achieve the lowest MAE/MSE on JHU-Crowd++, UCF-QNRF, and ShanghaiTech B. As for ShanghaiTech A, our loss achieves the second-best performance in MAE. 
    
    \subsection{Comparison with state-of-the-art methods}
    Table~\ref{tab:all-methods} shows the comparison between our loss and current SOTA. 
    Despite the fact that our method is simple, when combined with other methods, it achieves results that have a stable decrease in MAE and MSE compared with 
    the original methods.
    Compared to all previous methods, our loss combined with ConvNeXtS has a lower MAE and MSE on JHU-Crowd++, UCF-QNRF, and ShanghaiTech B than previous works in the table. Although our counting performance on ShanghaiTech A cannot exceed the performance of P2PNet~\cite{P2PNet} which is a purely point-based framework, our method is faster than P2PNet and saves more memory in the training phase.

    \subsection{Ablation study}
\begin{figure}[tp]
    \centering
    \includegraphics[width=\linewidth]{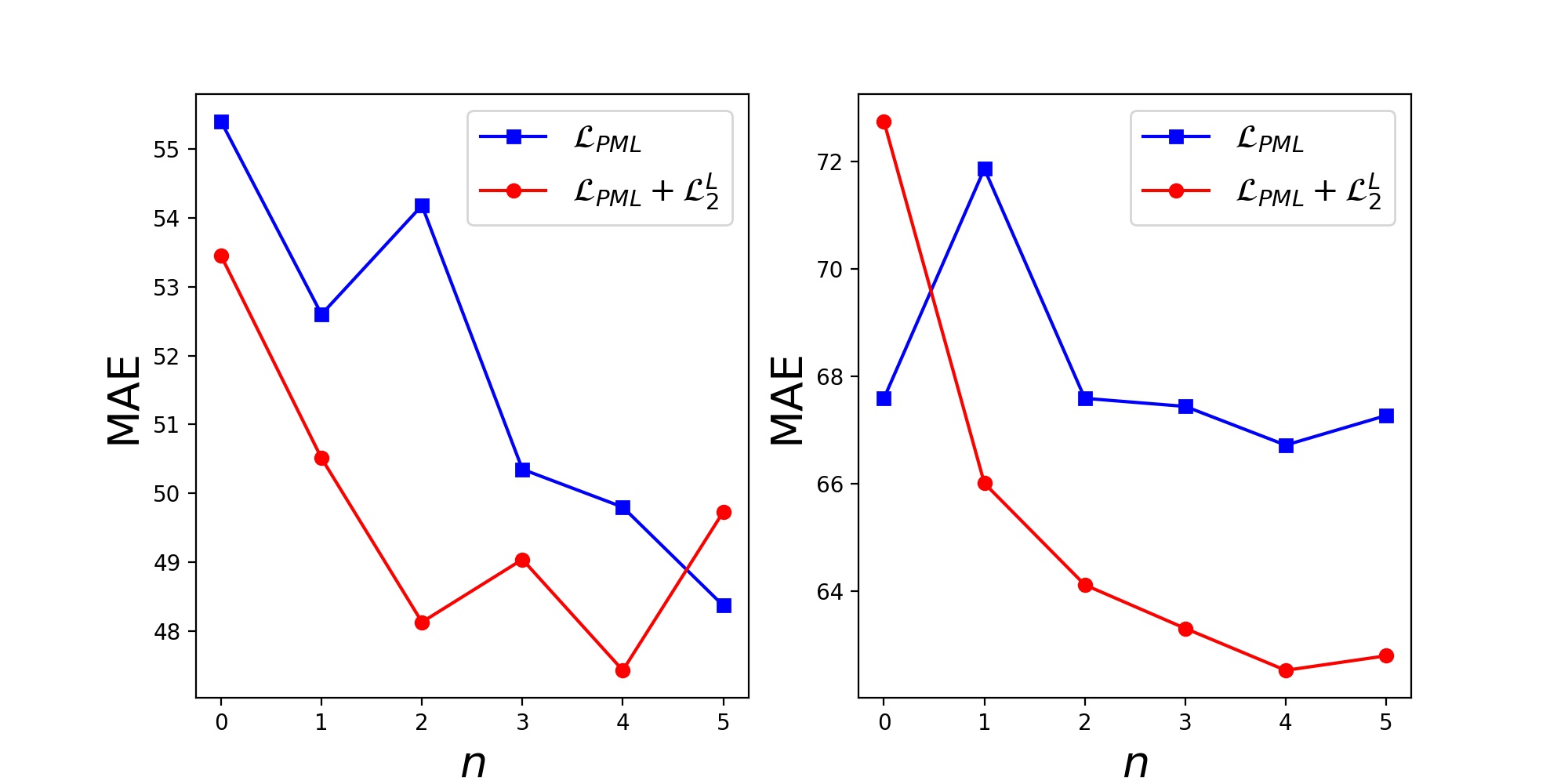}
    \caption{Validation MAE (left) and test MAE (right) of $\mathcal{L}_{PML}$ (blue curve) and $\mathcal{L}_{PML}+\mathcal{L}_2^L$ (red curve) on JHU-Crowd++. 
        $\mathcal{L}_{PML}$ and $\mathcal{L}_{PML}+\mathcal{L}_2^L$ are from Eq.~(\ref{loss-pml}) and Eq.~(\ref{loss-final}) respectively.
        The curves show the counting performances for $n$ with CSRNet as backbone. 
    }
    \label{fig:val_ab}
    \end{figure}

    Selecting appropriate resolutions for our loss is critical to 
    counting performance. 
    Our ablation experiments aim to verify whether the regularization is effective and what an appropriate $\mathbb{N}$ should be.
    As proved in Thm.~\ref{thm:N}, we only need to consider the situation that $\mathbb{N} = \{0, 1, \cdots, n\} \cup \{L\}$. Fig.~\ref{fig:val_ab} shows the performance on JHU-Crowd++, with $n$ from zero to five. 
    The model used in the ablation experiments is Csrnet referred to in Sec.~\ref{network}.  
    We did ablation experiments using the L2 norm as a regularization term. The experimental results show that it can improve performance a little. 
    The regularization term can reduce the average validation MAE by 2.07 and test MAE values by 2.83 on the JHU-Crowd++.

    \section{Conclusions}
    In this paper, we propose a simple and efficient loss function called PML for crowd counting. Our loss function, derived from the marginal likelihood, is proved to fit the posterior distribution better than L2 loss theoretically. By taking advantage of the multi-resolution density map, PML can learn to use multi-scale information to improve counting performance. We have demonstrated our loss function on four datasets with three common backbones in crowd counting with competitive results. 

{\small
    \bibliographystyle{ieee_fullname}
    \bibliography{egbib}
}
\end{document}